\documentclass{article}

\usepackage{microtype}
\usepackage{graphicx}
\usepackage{subfigure}
\usepackage{booktabs}  
\usepackage{hyperref}
\usepackage{multirow}

 \usepackage[accepted]{icml2025}

\usepackage{amsmath}
\usepackage{amssymb}
\usepackage{mathtools}
\usepackage{amsthm}
\usepackage[capitalize,noabbrev]{cleveref}

\theoremstyle{plain}

\theoremstyle{definition}

\theoremstyle{remark}

\usepackage[textsize=tiny]{todonotes}

\icmltitlerunning{Submission and Formatting Instructions for ICML 2025}

\begin{document}

\twocolumn[
% \icmltitle{Cross-subject BCI using multi-task learning model with identifiable spikes and associative memory networks}
%\icmltitle{\textcolor{blue}{ISAM-MTL: Cross-subject multi-task learning model with \\ identifiable spiking associative memory}}
\icmltitle{ISAM-MTL: Cross-subject multi-task learning model with identifiable spikes and associative memory networks}

\icmlsetsymbol{equal}{*}

\begin{icmlauthorlist}
\icmlauthor{Junyan Li}{xxx}
\icmlauthor{Bin Hu}{xxx}
\icmlauthor{Zhi-Hong Guan}{yyy}
\end{icmlauthorlist}

\icmlaffiliation{xxx}{School of Future Technology, South China University of Technology, Guangzhou, China}
\icmlaffiliation{yyy}{School of Artificial Intelligence and Automation, Huazhong University of Science and Technology, Wuhan, China}

\icmlcorrespondingauthor{Bin Hu}{huu@scut.edu.cn}

\icmlkeywords{Machine Learning, ICML}

\vskip 0.3in
]



\printAffiliationsAndNotice{\icmlEqualContribution} % otherwise use the standard text.

\begin{abstract}
Cross-subject variability in EEG degrades performance of current deep learning models, limiting the development of brain-computer interface (BCI). This paper proposes ISAM-MTL, which is a multi-task learning (MTL) EEG classification model based on identifiable spiking (IS) representations and associative memory (AM) networks. The proposed model treats EEG classification of each subject as an independent task and leverages cross-subject data training to facilitate feature sharing across subjects. ISAM-MTL consists of a spiking feature extractor that captures shared features across subjects and a subject-specific bidirectional associative memory network that is trained by Hebbian learning for efficient and fast within-subject EEG classification. ISAM-MTL integrates learned spiking neural representations with bidirectional associative memory for cross-subject EEG classification. The model employs label-guided variational inference to construct identifiable spike representations, enhancing classification accuracy. Experimental results on two BCI Competition datasets demonstrate that ISAM-MTL improves the average accuracy of cross-subject EEG classification while reducing performance variability among subjects. The model further exhibits the characteristics of few-shot learning and identifiable neural activity beneath EEG, enabling rapid and interpretable calibration for BCI systems.
\end{abstract}

\section{Introduction}
Brain-Computer Interfaces (BCIs) enable direct control of external devices through human brain activity patterns without involving muscle movements \cite{1chaudhary2016brain}, providing an effective means of information exchange between the brain and physical devices. BCIs hold significant potential for applications in medical rehabilitation and neuroscience research \cite{2lebedev2017brain}. Electroencephalography (EEG) is widely used in BCI systems due to its non-invasive nature and ease of operation, making it a preferred method for acquiring neural activity data from the brain. A typical BCI system consists of five components: EEG acquisition, preprocessing, feature extraction, classification, and task execution \cite{3lotte2015electroencephalography}. Among various EEG-based BCIs, the motor imagery (MI) paradigm has garnered extensive attention due to its broad application in medical rehabilitation. Critical steps in the MI paradigm are feature extraction from EEG signals and their classification.

Current deep learning methods like convolution and transformer have achieved good classification accuracy for MI EEG signals, yet most of them focus on intra-subject EEG classification \cite{32lawhern2018eegnet,33altaheri2022physics}. BCI systems based on such models require complex calibration processes when used for new subjects, which may hinder their broad applications. Due to the differences between subjects and acquisition equipments, cross-subject EEG data exhibit significant variability \cite{11melnik2017systems}. The variability in EEG signals causes excessive data training for deep learning models, which are detrimental to rapid calibration for BCI systems. Moreover, the representations of neural activity changes over time \cite{4degenhart2020stabilization}, further limiting the generalization of models trained on data from specific subjects to new subjects. Deep generative models like variational auto-encoder (VAE) and its variants show the feasibility of extracting complex nonlinear representations beneath EEG \cite{8zhou2020learning,eeg2vec,vaegan}. In view of such considerations, here comes an important question as how to reconstruct neural population activities with deep learning models that are efficient for cross-subject EEG classification and are simultaneously identifiable for relating neural representations and MI tasks.

    \begin{figure*}[h]
    \centering
    \includegraphics[width=0.98\linewidth]{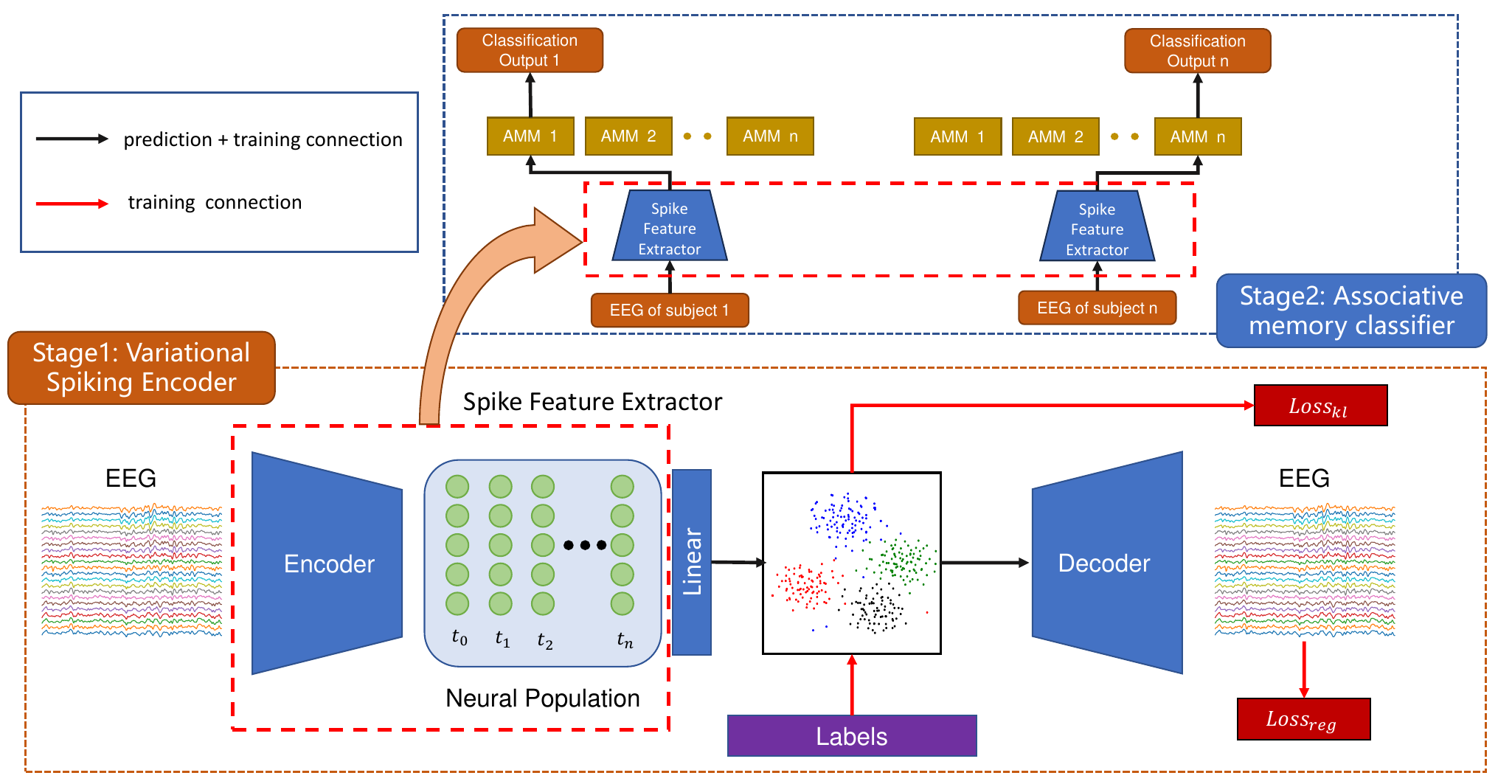}
    \caption{The architecture of the proposed ISAM-MTL model. The model proceeds with two stages. Stage 1: the cross-subject trained variational spiking encoder uses a 1D CNN and LIF neuron population to construct an autoencoder to extract low-dimensional spiking representations of EEG, and uses label-guided variational inference to enhance the identifiability of latent space spike. Stage 2: the associative memory classifier gives a AM trained within the subject using Hebbian learning for each subject, and the AM maps EEG to classes.}
    \label{fig:ISAM-MTL}
    \end{figure*}
    
Multi-task learning (MTL) leverages joint learning of neural activity features across subjects to increase the sample size, and may provide task-specific submodules to achieve high-accuracy EEG decoding across subjects or experiments \cite{10zheng2019multitask}. In contrast to single-task learning, MTL is an effective learning method that enhances the generalization of multiple related tasks, even when training samples for each subject task are limited. Moreover, low-dimensional representations of neural activity can effectively capture the high-dimensional structure of brain activity underlying motor control and reveal complex motor features \cite{5gallego2017neural}. Neural manifolds have been widely applied in both BCI systems and neuroscience research. Recent studies on neural manifolds have shown that the high-dimensional neural activity related to motor control exhibits similar structures across subjects \cite{6schneider2023learnable,7melbaum2022conserved}. Constructing separable and identifiable low-dimensional spikes could enhance the accuracy of EEG classification or regression analysis \cite{8zhou2020learning,9sani2024dissociative}. Neural populations in human hippocampus produce a hybrid mechanism of learning and associative memory, beneficial to few-shot training in classification tasks \cite{brain-inspired}. Overall, it suggests that neural activity, after manifold-based dimensionality reduction and associative memory, displays reduced variability and enhanced generalization across subjects, providing a doable approach for cross-subject EEG decoding. 

To address the issue of high variability in cross-subject EEG classification, this paper develops a multi-task learning model based on identifiable spiking low-dimensional representations and associative memory networks, abbreviated as ISAM-MTL (Figure \ref{fig:ISAM-MTL}). In the ISAM-MTL model, the classification of each subject's samples is treated as an subject task, enabling the extraction of shared features across different subjects' samples for cross-subject training and mapping these features to the respective sample categories. To adapt the model to new subjects with limited samples, an associative memory (AM) network is employed to build a classifier. The AM network, inspired by the principles of human brain memory, is an abstract framework that mimics the brain's ability to learn effective features from minimal demonstrations and perform accurate mappings \cite{15seitz2010sensory}. This associative memory mechanism contributes to the characteristic of rapid and few-shot learning of human brains. The pi-VAE framework is further borrowed to enhance the identifiability and separability of the latent spiking space \cite{8zhou2020learning}. By incorporating action labels as auxiliary variables, we can quantify the relationship between the latent variables of neural activity and task-specific variables. 
%Using variational inference, we construct disentangled and identifiable latent space distributions, thereby improving the model's classification performance. 
    
The main contributions of the paper are as follows. \vspace{-0.15in}
\begin{itemize}
\item A multi-task learning model, ISAM-MTL, is proposed for cross-subject EEG classification. ISAM-MTL integrates identifiable variational inference into spiking associative memory networks, guaranteeing separability of EEG spiking representations across subjects.The model has an average accuracy of 84.1\% and a standard deviation of 0.061 on the BCI Competition IV IIa dataset, representing the state-of-the-art performance for cross-subject classification.  \vspace{-0.1in}

\item Using labels as auxiliary variables to guide variational inference, ISAM-MTL enhances the identifiability of latent spiking representations, and reduces inter-subject EEG differences in the latent space through cross-subject pretraining. The distribution of different MI categories can be displayed by the model, through dimensionality reduction and visualization of the low-dimensional spiking representations. This identifiable characteristic helps to interpret the relationship between neural activity and MI tasks. \vspace{-0.1in}

\item Inspired by the hybrid learning and memory mechanism in human hippocampus, ISAM-MTL incorporates the spiking associative memory network to facilitate few-shot learning in conventional MTL. The model achieves an accuracy over 90\% with only 40 samples and enables to classify with as few as 2 class 5 shot, on the BCI Competition III Iva dataset. This few-shot characteristic is crucial to rapid calibration of real BCI systems, demonstrating the promise of our ISAM-MTL.
\end{itemize}

\section{Related Work}
Considering the cross-subject decoding accuracy, different deep learning models have been proposed for EEG classification and regression, such as transfer learning, domain adaptation, and contrastive learning. \citet{17wang2024tftl} presented TFTL, which incorporates a domain discriminator and uses resting-state data from target subjects to construct subject-specific domains, enabling transfer learning across individuals. \citet{18waytowich2016spectral} developed the spectral transfer information geometry (STIG) model, which leverages information geometry to rank and combine predictions from classifier ensembles, achieving unsupervised transfer learning for single-trial detection. \citet{19zhi2024supervised} proposed SCLDGN, which employs domain-invariant supervised contrastive learning with deep correlated alignment and class regularization blocks, achieving promising results across multiple motor imagery datasets. As a solution for cross-individual EEG decoding, multi-task learning (MTL) optimizes multiple loss functions simultaneously while sharing common features across different tasks, enabling the learning of more generalized representations. \citet{10zheng2019multitask} assigned a task to each subject's samples and developed an efficient algorithm based on regularized tensors. It employs the Fisher discriminant criterion for feature selection and uses the alternating direction method of multipliers for optimization. However, these deep learning models lack identifiable subject-invariant hidden state, resulting in poor cross-subject performance.

To enhance the separability of neural activity across different actions and reduce variability crossing subjects, it is a viable approach to mapping high-dimensional neural activity into identifiable low-dimensional representations. \citet{8zhou2020learning} proposed the Poisson identifiable VAE (pi-VAE), which uses label variables to construct conditionally independent Gaussian mixtures in the latent space, providing disentangled and identifiable latent representations. \citet{6schneider2023learnable} treated trial information as task-independent variables and obtained an embedding space that no longer carried this information, enabling the construction of neural manifolds without explicit modeling of the data generation process. These generative models demonstrate the feasibility of neural manifolds for decoding neural activity data across sensory and motor tasks, ranging from simple to complex behaviors. Nonetheless, manifold-based models often entail overly learned features and continuous latent spaces, which may constrain their applicability due to the limited samples in real BCI systems.

Brain science suggests that it is feasible to enhance the model generalization across subjects and tasks by incorporating Hebbian-based associative memory into deep learning. \citet{22miconi2018differentiable} combined associative memory following Hebbian rules with traditional backpropagation neural networks, enabling efficient learning of small-sample image datasets. \citet{23wu2022brain} applied a similar structure to spiking neural networks, where synaptic weights are updated through a combination of backpropagation-based global learning and local Hebbian learning. This hybrid global-local approach performed well in various tasks like fault-tolerant learning, few-shot learning, and continual learning. From the best of our knowledge, little effort has been laid on developing cross-subject models based on associative memory for decoding EEG signals, with emphasis on identifying task-specific neural activity and generalizing to few-shot learning scenario.

    \begin{figure}[h]
        \centering
        \includegraphics[width=0.98\linewidth]{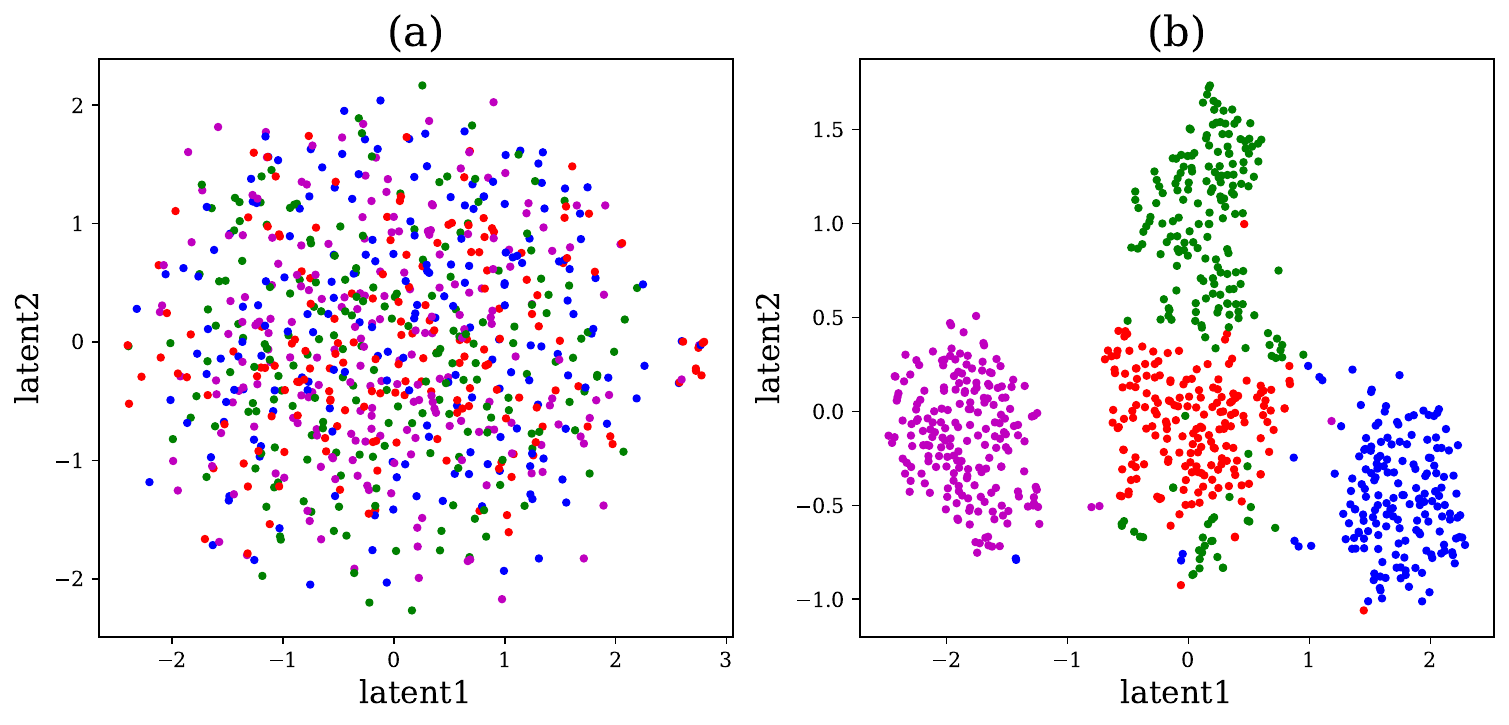}
        \caption{The distribution of latent spike features after dimensionality reduction using t-SNE. (a) Variational inference without label guidance. (b) Variational inference with label guidance.}
        \label{fig:latent}
    \end{figure}

\section{Method}
Figure \ref{fig:ISAM-MTL} presents the computational framework of the ISAM-MTL model. Fed with EEG signals, ISAM-MTL consists of a variational spiking encoder and an associative memory classifier. Two stages are included for the training and prediction process of the model.
    
{\bf Stage 1}: The spiking encoder uses a 1D convolutional neural network (CNN) combined with a spiking neural network to encode EEG signals into low-dimensional spiking representations. Label-guided variational inference is then applied to map these representations to a latent space distribution. Subsequently, a 1D convolutional decoder reconstructs the EEG signals. Here, the model uses labels as auxiliary variables to enhance the identifiability of the latent space (Figure \ref{fig:latent}).

{\bf Stage 2}: The associative memory classifier is employed for multi-subject classification. It consists of multiple associative memory matrices (AMM), with each matrix corresponding to the classification task of a specific subject. The classifier maps the low-dimensional spiking representations to their respective categories. During training, the parameters of the spiking encoder are frozen, and the associative memory matrices are trained for each task. Here, the input and category labels are represented as input-output pattern pairs $\{\mathbf{x}_i,\mathbf{y}_i\}$, where $\mathbf{x}_i\in R^{nt}$ is the low-dimensional spiking representation vector passed through the spiking encoder, with $n$ representing the number of neurons in the population and $t$ is the length of the reduced-dimensional spiking time series. $\mathbf{y}_i$ is the one-hot encoded target vector of the labels.

The associative memory matrix follows Hebbian learning to form hetero-associative memory, matching the input patterns to the corresponding output patterns. The training of the associative memory classifier is a linear time complexity algorithm, which can achieve efficient learning.
    
\subsection{Spike Feature Extractor}
As shown in Figure \ref{fig:ISAM-MTL}, the spike feature extractor includes an encoder $E$ and decoder $D$ built with 1D convolutional layers and a neural population. Each convolutional layer in the encoder uses a kernel length of 5 and is activated by the ReLU function. The input signal $x\in R^{n\times t}$ is downsampled by two 1D max-pooling layers within the encoder to $1/4$ of its original length, producing the hidden signal. The encoded signal is then directly input into the spiking neurons as current through the fully connected layer, recording the spike sequence emitted by the neuron and converting it into a spike vector $Spk\in R^{nt/4}$. Here, we use the leaky integrate-fire (LIF) neuron model \cite{dendritic,adaptive}. The membrane potential $v^t$ and spike $s^t$ changes with time $t$ as
     \begin{equation}
        \label{eq:LIF}
            \begin{aligned}
                v^t &= (1-\tau)v^{t-1}-s^{t-1}v_{th}+\sum_{i}(w_iI_i^{t-1}),\\
                s^t &= step(v^t-v_{th}),
            \end{aligned}
    \end{equation}
where $\tau$ is the time constant and $I$ is the input current, and $step(x) = 
            \begin{cases}
                1,&x\ge0 \\
                0,&x<0
            \end{cases}$. When the potential reaches the spike threshold $v_{th}$, the neuron fires a spike and returns to the potential $0$. The all-or-none nature of spike makes the backpropagation of gradients difficult \cite{adaptive}. Here, we use the $rect(x) = 
            \begin{cases}
                1,&|x|\le0.5 \\
                0,&|x|>0.5
            \end{cases}$ as a surrogate gradient to the $step$ function. 

In the decoder stage, the spike vector undergoes variational inference and is then upsampled to its original length by two 1D transposed convolutions.

\subsection{Identifiable Variational Inference}
To make the latent space identifiable, we use a fully connected layer to map the spike vector $Spk\in R^{nt/4}$ to the latent space distribution $p(z|x)$. Referring to the variational inference process of pi-VAE \cite{8zhou2020learning}, we use the label as an auxiliary variable $u$ to guide the identifiability of the latent space (Figure \ref{fig:latent}). The variational inference process guided by auxiliary variables is
    $$
    p_{\theta}(x,z|u) = p_f(x|z)p_{\lambda}(z|u),
    $$
where label prior $p_{\lambda}(z|u)$ is conditionally independent Gaussian distribution. The model learns an approximate posterior $q(z|x,u)$ and the true posterior $p(z|x,u)$ by maximizing the evidence lower bound (ELBO). It follows that 
    $$
    \mathop{\arg\min}\limits_{\theta,\lambda} KL[q(z|x,u)||p(z|x,u)],
    $$
where
    $$
    KL[q(z|x,u)||p(z|x,u)] = \int_zq(z|x,u)log\frac{q(z|x,u)}{p(z|x,u)}dz.
    $$
According to the Bayesian formula, $p(z|x,u)=\frac{p(x,z|u)}{p(x|u)}$. One may have  
    $$
    logp(x|u)=-KL[q(z|x,u)||p(z|x,u)]-ELBO,
    $$
where
    $$
    ELBO = E_{q(z|x,u)}[log(p(x,z|u))-log(q(z|x,u))].
    $$
In order to minmize $KL[q(z|x,u)||p(z|x,u)]$, We optimize the spike feature extractor parameters $\theta$ and the decoder parameters $\lambda$ to maximize the ELBO,
    $$
    \mathop{\arg\max}\limits_{\theta,\lambda}E_{q(z|x,u)}[log(p(x,z|u))-log(q(z|x,u))].
    $$
After learning $q(z|x,u)$, the decoder adopts a structure similar to the convolutional encoder in the spiking feature extractor, uses two 1D transposed convolutions to up-sample the data, and reconstructs the variable $z$ in the latent space to the original EEG signal.

\subsection{Associative Memory Classifier}
In order to ensure few-shot learning in cross-subject EEG classification,  we build associative memory networks to map spiking neural activity to MI labels. For each subject’s EEG classification, it is treated as one single task and is assigned with an associative memory matrix (AMM). Let the input-output pattern pair be $\{\mathbf{x},\mathbf{y}\}$, where $\mathbf{x}\in R^n$ is the input spike vector and $\mathbf{y}\in R^m$ is output one-hot vector. In the memory retrieval stage, the iterative process of the input-output pattern pair $\{\mathbf{x},\mathbf{y}\}$ using the AMM $\mathbf{W}$ is
    \begin{equation}
        \label{eq:BAM}
            \begin{aligned}
                \mathbf{y}^{t+1} &= sgn(\mathbf{W}\mathbf{x}^t), \\
                \mathbf{x}^{t+1} &= sgn(\mathbf{W}^T\mathbf{y}^t),
            \end{aligned}
        \end{equation}
where $sgn=\begin{cases}
            -1&x\le0 \\
            +1&x>0
        \end{cases}.$ When the associative memory system is stable, $x^{t+1}=x^t, y^{t+1}=y^t$, and the classification result is $label = \mathop{\arg\max}\mathbf{Wx}$.

The established AMMs are trained with Hebbian learning, dictating the learning rule for hippocampal neuronal connections \cite{27kelso1986hebbian}. That is, when two neurons fire together, their connection strengthens, otherwise, the connection weakens. The associative memory matrix $\mathbf{W}_k$ of the task $k$ is
    $$
    \mathbf{W}_k = \sum_j\mathbf{y}_k^j\mathbf{x}_k^{jT}.
    $$
The time complexity of the Hebbian learning algorithm under parallel computing conditions is $\mathcal{O}(n)$, where $n$ is the number of samples. This means that the model can adapt to new subjects with higher efficiency in the second stage. 

\section{Experiments}
In this section, we apply ISAM-MLT to decode the MI paradigm, which uses the EEG classification model to classify the EEG signals into different actions. Here we utilize the BCI Competition datasets and achieve an average accuracy over 84\% on the BCI Competition IV IIa subset, surpassing the current state-of-the-art (SOTA).

\begin{table*}[h]
\caption{Cross-subject classification comparison experiment on BCI Competition IV IIa dataset\label{tab:experiment}}
\centering
\begin{tabular}{c|c|ccccc}
\hline\hline
\multicolumn{2}{c|}{Models}  & Shallow-MDTL & TFTL & DS-KTL & Tensor-based MTL  & Ours  \\ \hline
\multicolumn{2}{c|}{Sources}&\cite{30li2023mdtl}&\cite{17wang2024tftl}&\cite{31luo2023dual}&\cite{10zheng2019multitask}& /\\ \hline
\multirow{9}{*}{Subjects}&1    & \textbf{0.847}                & 0.826        & 0.66           & 0.84                  & 0.808 \\
&2    & 0.618                & 0.674        & 0.458          & 0.573                 & \textbf{0.890}  \\
&3    & 0.861                & \textbf{0.951}        & 0.819          & 0.549                 & 0.853 \\
&4    & 0.688                & 0.799        & 0.694          & \textbf{0.959}                 & 0.846 \\
&5    & 0.729                & 0.743        & 0.646          & \textbf{0.912}                 & 0.825 \\
&6    & 0.667                & 0.757        & 0.486          & \textbf{0.826}                 & 0.717 \\
&7    & 0.806                & 0.736        & 0.785          & 0.792                 & \textbf{0.825} \\
&8    & 0.757                & 0.882        & 0.951          & 0.835                 & \textbf{0.958} \\
&9    & 0.792                & \textbf{0.924}        & 0.813          & 0.819                 & 0.842 \\ \hline
&avg  & 0.750                & 0.810        & 0.701          & 0.790                 & \textbf{0.841} \\
&std  & 0.766                & 0.088        & 0.161          & 0.131                 & \textbf{0.061} \\ \hline \hline
\end{tabular}
\end{table*}
    
\subsection{Dataset Description}

\paragraph{BCI Competition III Iva} It is a binary classification dataset \cite{28dornhege2004boosting}, including right-hand and foot movement imagery tasks performed by 5 subjects. Each task includes 118 channels of EEG signals obtained at a sampling rate of 100 Hz within 3 seconds.

\paragraph{BCI Competition IV IIa} It is a four-category dataset \cite{29brunner2008bci}, including motor imagery tasks of the left hand, right hand, feet, and tongue performed by 9 subjects. Each task includes 22 channels of EEG signals and 3 channels of EOG signals obtained at a sampling rate of 250 Hz within 3 seconds.
    
    \begin{figure}[bt]
        \centering
        \includegraphics[width=0.98\linewidth]{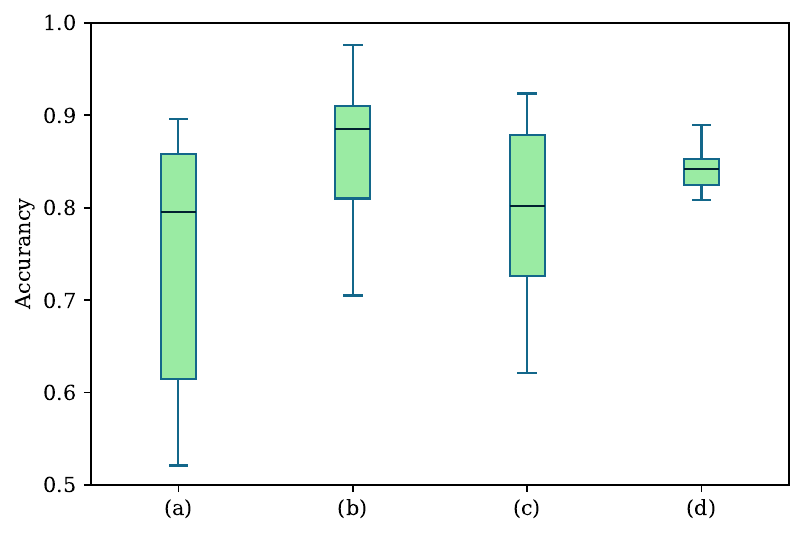}
        \caption{Comparison of ISAM-MTL with other subject-specific classification models on the BCI Competition IV IIa dataset, (a) EEGNet\cite{32lawhern2018eegnet}, (b) ATCNet\cite{33altaheri2022physics} (SOTA), (c) FBMSNet\cite{34liu2022fbmsnet}, (d) ISAM-MTL.}
        \label{fig:in_sunject}
    \end{figure}

\subsection{Performance Evaluation}
\paragraph{Comparative studies} The comparison between ISAM-MTL and other cross-subject classification models is summarized in Table \ref{tab:experiment}. On the BCI Competition IV IIa dataset, our model achieves the SOTA cross-subject average accuracy. Figure \ref{fig:in_sunject} compares ISAM-MTL with other classical subject-specific classification models on the same dataset. While ISAM-MTL uses cross-subject training and testing, the other models are trained and tested using subject-specific data. The results show that the accuracy of our model on cross-subject data is comparable to the SOTA models for subject-specific classification.Compared to other cross-subject classification models, ISAM-MTL achieves the smallest standard deviation in accuracy across different samples, demonstrating that our model provides stable classification performance for various individuals. Additionally, when expanding to new tasks, only the associative memory matrices need to be retrained. This retraining process leverages highly efficient Hebbian-like learning, making our model highly scalable.

\begin{figure}[bt]
        \centering
        \includegraphics[width=0.98\linewidth]{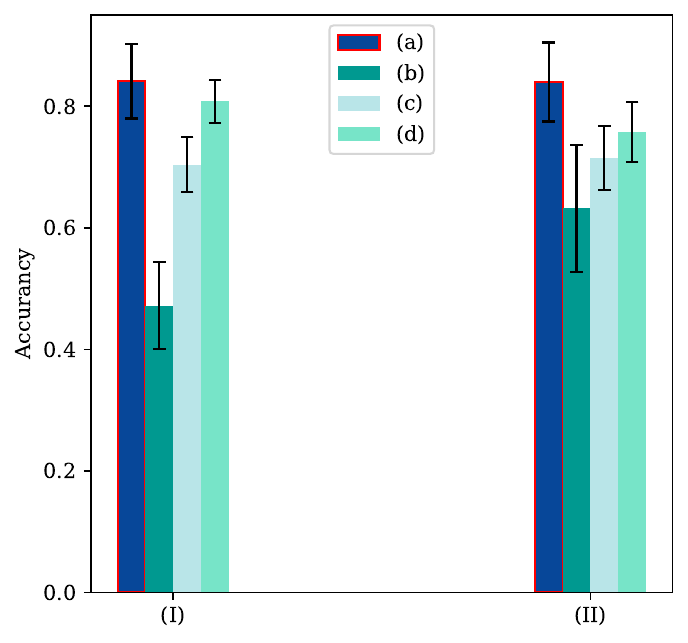}
        \caption{Ablation experiments of our model on (I) BCI Competition IV IIa and (II) BCI Competition III Iva. (a) ISAM-MTL. (b) Remove label-guided variational inference. (c) Gradient descent fully connected layers replace associative memory matrix model. (d) Tanh activation function replaces LIF neurons.}
        \label{fig:ablation}
    \end{figure}

\paragraph{Ablation studies} We conduct ablation experiments on two datasets, as shown in Figure \ref{fig:ablation}. On both datasets, the ISAM-MTL model outperforms other simplified models. The experimental results demonstrate that the components of the model, including label-guided variational inference, associative memory networks, and the LIF spiking neuron model, all significantly contribute to the model's performance.

\subsection{Few-Shot Learning}
\begin{figure}[bt]
        \centering
        \includegraphics[width=0.98\linewidth]{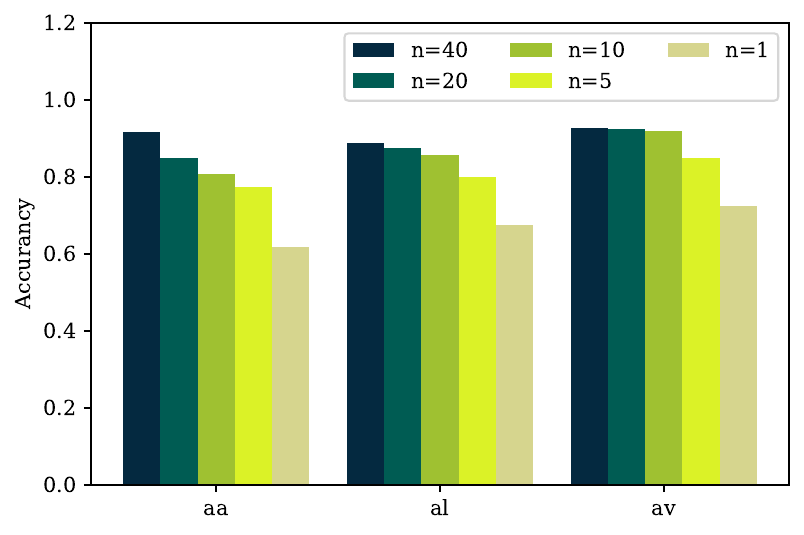}
        \caption{ISAM-MTL few-shot learning on the BCI Competition III Iva dataset, where $n$ is the number of samples for each class.}
        \label{fig:few_shot}
    \end{figure}

We conduct cross-subject few-shot learning experiments on subjects aa, al, and av, selected from the BCI Competition III Iva dataset. The spiking feature extractor is trained using the full training set, while the associative memory matrix is computed using a reduced training set. Figure \ref{fig:few_shot} shows the classification accuracy of ISAM-MTL under five scenarios of few-shot learning, on the BCI Competition III Iva dataset. As can be seen, the ISAM-MTL model maintains relatively stable performance as the sample size per class decreases and is capable of supporting as few as 5 samples for 2-class learning. Additionally, the model achieves an average accuracy of 65.4\% for single-sample learning. 

\subsection{Neural Activity Identification}
To demonstrate the identifiability of latent spiking representations, we compare the latent spaces learned by the ISAM-MTL model with that obtained from a spiking feature encoder without label guidance using the BCI Competition IV IIa dataset. As indicated by Figure \ref{fig:latent}, label-guided variational inference enhances the identifiability of the low-dimensional spiking features across four motor imagery (MI) classes.

    \begin{figure*}[bt]
    \centering
    \includegraphics[width=0.98\linewidth]{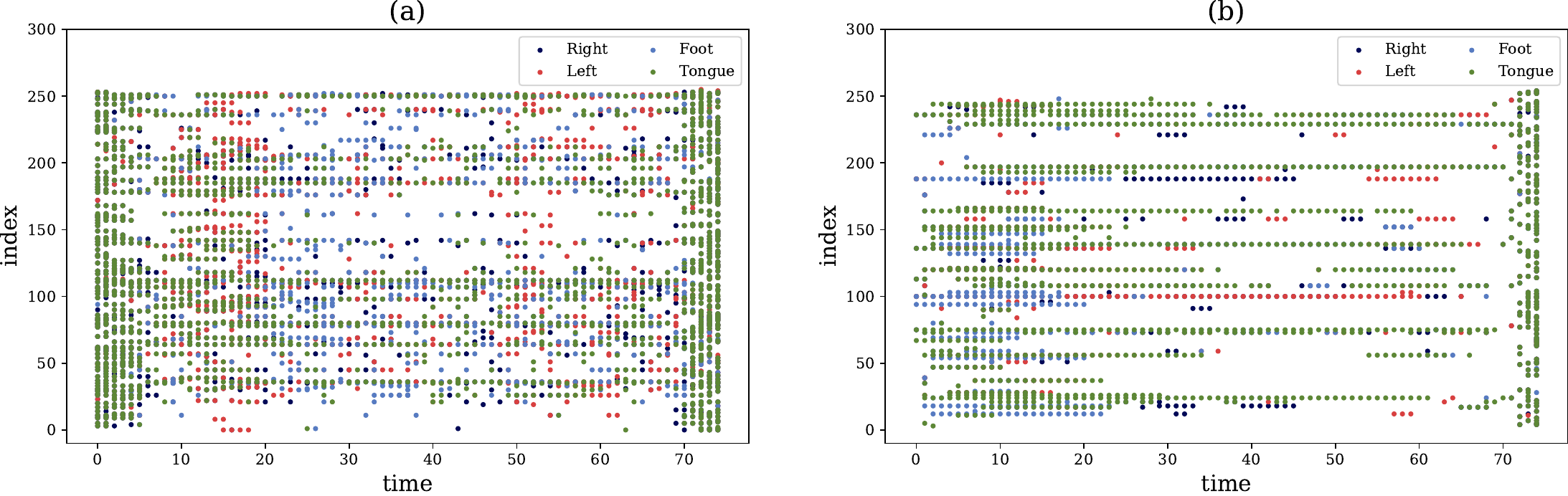}
    \caption{Characteristic spike firing diagram of hidden-layer neurons as per four types of motor imagery tasks, on the BCI Competition IV IIa dataset. (a) Model without label-guided variational inference. (b) ISAM-MTL.}
    \label{fig:latent_spike}
    \end{figure*}

We further analyze the motor imagery spiking features $\hat{\mathbf{x}}$ by passing the one-hot encoded MI class labels $\mathbf{y}$ through the associative memory matrix (AMM) in reverse. It is calculated by $\hat{\mathbf{x}}=\mathbf{W}^T\mathbf{y}$. Figure \ref{fig:latent_spike} depicts the characteristic spike firing of neurons in the hidden layer. It is shown that ISAM-MTL improves the specificity of spiking neurons' responses to particular motor classes through the auxiliary variable mechanism. By visualizing the latent spiking representations, this experiment further enhances the interpretability of the latent space, providing a reliable basis for the model’s accurate classification.

\section{Conclusion}
In this paper, we propose ISAM-MTL, a multi-task learning model leveraging identifiable spiking representations and associative memory networks for cross-subject motion imagery EEG classification. The established spiking feature extractor uses 1D convolutional layers and LIF neurons to transform EEG into low-dimensional spiking representations. To ensure the identifiability of the high-dimensional spiking representations, we adopt the variational inference from pi-VAE, constructing a mixture of Gaussian distributions in the latent space through label-guided auxiliary variables. Using a multi-task classifier based on associative memory networks, ISAM-MTL employs Hebbian learning with linear time complexity to quickly map latent spiking representations of EEG signals to categories. Additionally, the model is capable of learning accurate mappings with a small number of samples.

Low-dimensional spiking representations of the latent space provide possible support for interpretability of neural activity under EEG. However, the specific information encoded by each LIF neuron and its spike timing remain unclear, which should be one focus of our future work. Sparse connections can improve the energy-efficient performance of spiking neural networks \cite{dendritic}. Another future work would be building sparse spiking associative memory networks using pruning operations and testing the to-be-optimized model with application to BCI systems.

%\section*{Accessibility}
%Authors are kindly asked to make their submissions as accessible as possible for everyone including people with disabilities and sensory or neurological differences.
%Tips of how to achieve this and what to pay attention to will be provided on the conference website \url{http://icml.cc/}.

%\section*{Software and Data}

%If a paper is accepted, we strongly encourage the publication of software and data with the camera-ready version of the paper whenever appropriate. This can be done by including a URL in the camera-ready copy. However, \textbf{do not} include URLs that reveal your institution or identity in your submission for review. Instead, provide an anonymous URL or upload the material as ``Supplementary Material'' into the OpenReview reviewing system. Note that reviewers are not required to look at this material when writing their review. 

%\section*{Acknowledgements}

%\textbf{Do not} include acknowledgements in the initial version of the paper submitted for blind review.

%If a paper is accepted, the final camera-ready version can (and usually should) include acknowledgements.  Such acknowledgements should be placed at the end of the section, in an unnumbered section that does not count towards the paper page limit. Typically, this will include thanks to reviewers who gave useful comments, to colleagues who contributed to the ideas, and to funding agencies and corporate sponsors that provided financial support.

%\section*{Impact Statement}

\bibliography{icml2025_ref}
\bibliographystyle{icml2025}

%%%%%%%%%%%%%%%%%%%%%%%%%%%%%%%%%%%%%%%%%%%%%%%%%%%%%%%%%%%%%%%%%%%%%%%%%%%%%%%
%%%%%%%%%%%%%%%%%%%%%%%%%%%%%%%%%%%%%%%%%%%%%%%%%%%%%%%%%%%%%%%%%%%%%%%%%%%%%%%
% APPENDIX
%%%%%%%%%%%%%%%%%%%%%%%%%%%%%%%%%%%%%%%%%%%%%%%%%%%%%%%%%%%%%%%%%%%%%%%%%%%%%%%
%%%%%%%%%%%%%%%%%%%%%%%%%%%%%%%%%%%%%%%%%%%%%%%%%%%%%%%%%%%%%%%%%%%%%%%%%%%%%%%
%\newpage
%\appendix
%\onecolumn
%\section{Detailed Setting of The Experiments}

%You can have as much text here as you want. The main body must be at most $8$ pages long.
%For the final version, one more page can be added.
%If you want, you can use an appendix like this one.  

%The $\mathtt{\backslash onecolumn}$ command above can be kept in place if you prefer a one-column appendix, or can be removed if you prefer a two-column appendix.  Apart from this possible change, the style (font size, spacing, margins, page numbering, etc.) should be kept the same as the main body.
%%%%%%%%%%%%%%%%%%%%%%%%%%%%%%%%%%%%%%%%%%%%%%%%%%%%%%%%%%%%%%%%%%%%%%%%%%%%%%%
%%%%%%%%%%%%%%%%%%%%%%%%%%%%%%%%%%%%%%%%%%%%%%%%%%%%%%%%%%%%%%%%%%%%%%%%%%%%%%%
\end{document}